\documentclass[utf8]{FrontiersinHarvard} 

\usepackage{url,hyperref,lineno,microtype,subcaption}
\usepackage[onehalfspacing]{setspace}
\usepackage{booktabs}
\usepackage{multirow}
\usepackage{algorithm}
\usepackage{comment}
\usepackage{algorithmic}
\usepackage{textcomp}

\def\keyFont{\fontsize{8}{11}\helveticabold }
\def\firstAuthorLast{Khan {et~al.}} 
\def\Authors{Ahmad Saeed Khan\,$^{1,*}$, Erik Schaffernicht\,$^{1}$ and Johannes Andreas Stork\,$^{1}$}
\def\Address{$^{1}$Örebro University, Sweden }
\def\corrAuthor{Ahmad Saeed Khan}

\def\corrEmail{ahmad-saeed.khan@oru.se}

\usepackage{todonotes}
\presetkeys%
{todonotes}%
{inline,backgroundcolor=yellow}{}

\begin{document}
\onecolumn
\firstpage{1}

\title[DFW]{DFW: A Novel Weighting Scheme for Covariate Balancing and Treatment Effect Estimation} 

\author[\firstAuthorLast ]{\Authors} 
\address{} 
\correspondance{} 

\extraAuth{}

\maketitle

\begin{abstract}
	Estimating causal effects from observational data is challenging due to selection bias, which leads to imbalanced covariate distributions across treatment groups. Propensity score-based weighting methods are widely used to address this issue by reweighting samples to simulate a randomized controlled trial (RCT). However, the effectiveness of these methods heavily depends on the observed data and the accuracy of the propensity score estimator. For example, inverse propensity weighting (IPW) assigns weights based on the inverse of the propensity score, which can lead to instable weights when propensity scores have high variance—either due to data or model misspecification—ultimately degrading the ability of handling selection bias and treatment effect estimation. To overcome these limitations, we propose Deconfounding Factor Weighting (DFW), a novel propensity score-based approach that leverages the deconfounding factor—to construct stable and effective sample weights. DFW prioritizes less confounded samples while mitigating the influence of highly confounded ones, producing a pseudopopulation that better approximates a RCT. Our approach ensures bounded weights, lower variance, and improved covariate balance.While DFW is formulated for binary treatments, it naturally extends to multi-treatment settings, as the deconfounding factor is computed based on the estimated probability of the treatment actually received by each sample. Through extensive experiments on real-world benchmark and synthetic datasets, we demonstrate that DFW outperforms existing methods, including IPW and CBPS, in both covariate balancing and treatment effect estimation.
	\tiny
	\keyFont{ \section{Keywords:} propensity score, weighting, confounding, covariate balancing, treatment effect }
\end{abstract}

\section{Introduction}

Answering causal inference questions—such as quantifying the potential effectiveness of a specific drug on a given health condition—is more reliable when we have access to Randomized Controlled Trial (RCT) data \citep{Imbens}. In RCTs, the distribution of covariates is balanced across treatment groups, meaning that treatment assignment is independent of subject characteristics, which makes estimating treatment effects straightforward. However, conducting RCTs is often infeasible or unethical \citep{JudeaP}. As a result, we rely on observational studies, where covariate-driven treatment assignment introduces selection bias and leads to systematic covariate imbalance between treatment groups.

To mitigate such bias, propensity score-based weighting techniques are employed to emulate the balance achieved in RCTs \citep{PRrose,IPW,JudeaP,Imbens}.The \emph{propensity score} is defined as the conditional probability of receiving treatment given observed covariates:
	\[
	e(\mathbf{x}) = P(t = 1 \mid \mathcal{X} = \mathbf{x}),
	\]
	where \(t \in \{0,1\}\) denotes the binary treatment assignment and \(\mathcal{X}\) is the set of observed covariates. These methods assign weights to samples such that the reweighted treatment and control groups become statistically comparable, reducing confounding bias in downstream treatment effect estimation.

A central limitation of weighting-based methods lies in their reliance on estimated propensity scores, which are not observed but must be inferred from the data. Consequently, their effectiveness is highly sensitive to the type of data and the accuracy of the propensity score model, with no performance guarantees across diverse settings. These methods often fail when there are strong differences in the distribution of covariates between the groups being compared (i.e., high selection bias), or when the data lacks enough similar samples across treatment groups to allow for fair comparisons (i.e., limited overlap). Furthermore, \citet{Olmos2015} highlight the susceptibility of treatment effect estimates to model specification. Their empirical analysis on the Jobs dataset demonstrates that varying the model used for propensity score estimation can lead to substantial shifts in the estimated treatment effect—ranging from statistically significant to non-significant—underscoring the instability introduced by model misspecification \citep{Harder2010PropensityST}.

For instance, Inverse probability weighting (IPW) is widely regarded as a gold-standard method for balancing covariate distributions across treatment groups using propensity scores. By weighting each sample by the inverse of propensity score, IPW effectively mitigates selection bias \citep{PRrose}. However, IPW weights are unbounded. When covariate distributions differ between treatment groups, some individuals may have an extremely low probability of receiving treatment. Inverting such propensity scores leads to extreme weights, inflating their variance (see Fig. \ref{fig:1}) and resulting in suboptimal balance and inaccurate treatment effect estimation \citep{Cole2008ConstructingIP}. IPW inherently suffers from undesirable mathematical properties, notably the potential for high-variance weights, as explained in Subsection~\ref{sec:proof}. In real-world settings, factors such as high covariate imbalance and misidentification of propensity score model can severely compromise the robustness of IPW \citep{Olmos2015}, as illustrated in Figure~\ref{fig:1} and Table~\ref{tab:1}. This highlights the need for a weighting scheme that is robust to such practical challenges and less sensitive to the type of underlying data or the choice of estimation model. Moreover, such a scheme should also exhibit desirable mathematical properties that directly address IPW’s limitations, including unboundedness and high variance in weights.

To address key limitations of existing methods—such as high-variance weights and sensitivity to propensity score model misspecification—we introduce Deconfounding Factor Weighting (DFW), a propensity score-based approach that adjusts sample weights using the deconfounding factor, defined as 1\textminus the probability of receiving the observed treatment. DFW constructs a pseudopopulation that more closely approximates the covariate distribution of a randomized controlled trial (RCT), enabling reliable and unbiased causal comparisons. Importantly, DFW retains desirable mathematical properties, including boundedness and reduced variance in the weights, directly addressing the instability issues inherent in IPW. In particular, DFW intuitively avoids the inflation of weights common in IPW (see intuition in Subsection~\ref{sec:proof}), contributing to improved performance, as demonstrated in Table~\ref{tab:1}, and Figure~\ref{fig:1}. Overall, DFW offers robust and stable treatment effect estimation, even under challenging conditions such as limited covariate overlap and high selection bias.

We evaluate DFW using established statistical balancing criteria as well as treatment effect estimation, and our results demonstrate that DFW outperforms commonly used baselines, including IPW, covariate balancing propensity score (CBPS) and Overlap weighting on two real-world benchmarks and synthetic datasets.

\subsection{Our Contributions}

\begin{itemize} 
	\item We propose a novel deconfounding factor–based weighting scheme that effectively balances covariate distributions and improves treatment effect estimation with stable and low variance weights. 
	\item We develop a computationally efficient and scalable approach suitable for large-scale datasets encompassing both binary and multiple treatment scenarios.
	\item We demonstrate through extensive experiments on benchmark and synthetic datasets that our method consistently outperforms standard baselines in both covariate balancing and treatment effect estimation based on established metrics and evaluation criteria. 
\end{itemize}

\section{Background}
In observational studies, treatment assignment is not randomized, leading to selection bias—systematic differences in the covariate distributions between treatment groups—which results in imbalance that can distort causal effect estimates. For instance, if older patients—who naturally have different outcomes—are more likely to receive a new drug, the observed effect may be due to age rather than the drug itself. In this case, age is a confounding variable affecting both treatment assignment and outcome, and failing to adjust for it can result in misleading conclusions \citep{PRrose,Anpeng}. A widely used tool to mitigate this bias is the \emph{propensity score}, defined as the probability of receiving treatment conditional on observed covariates \citep{PRrose}:
\[
e(\mathbf{x}) = P(t = 1 \mid \mathcal{X} = \mathbf{x}).
\]
When conditioning on the propensity score, the distribution of covariates should ideally be balanced across treatment groups, enabling unbiased estimation of treatment effects.

\textbf{Inverse Propensity Weighting (IPW)} is a foundational method that reweights samples to mimic a randomized trial \citep{PRrose,IPW}. Each sample is weighted as follows:
\[
w = \frac{t}{e(\mathbf{x})} + \frac{1 - t}{1 - e(\mathbf{x})},
\]
where \( t \in \{0, 1\} \) denotes observed treatment. 

\textbf{Covariate Balancing Propensity Score (CBPS)} \citep{Imai} \citep{Christian} CBPS reframes propensity score estimation as a dual objective problem: it simultaneously fits the propensity model and ensures balance between covariates across treatment groups. The objective function explicitly minimizes the imbalance in the weighted covariate means between treated and control groups:

\[
\min_{\beta} \left\| \frac{1}{n} \sum_{i=1}^n \left( \frac{t_i \mathbf{x}_i}{e(\mathbf{x}_i; \beta)} - \frac{(1 - t_i) \mathbf{x}_i}{1 - e(\mathbf{x}_i; \beta)} \right) \right\|^2,
\]
where \( t_i \in \{0,1\} \) is the observed treatment, \( \mathbf{x}_i \in \mathbb{R}^d \) are the observed covariates,
\( e(\mathbf{x}_i; \boldsymbol{\beta}) = P(t_i = 1 \mid \mathbf{x}_i) \) is the propensity score model, typically parameterized using logistic regression and $(\|\cdot\|)$ is the Euclidean norm.

\textbf{Stratification} divides data into strata (blocks) based on estimated propensity scores \citep{Imbens}. Within each stratum, covariates are assumed to be approximately balanced. The average treatment effect (ATE) is then calculated as a weighted average across strata:
\[
\mathrm{ATE} = \sum_{j=1}^{J} q(j) \left[ \bar{Y}_{t}(j) - \bar{Y}_{c}(j) \right],
\]
$\bar{Y}_{t}(j)$ and $\bar{Y}_{c}(j)$ denote the mean outcomes for treated and control subjects in block $j$ and $q(j)$ represents the proportion of the total sample contained in that block \citep{Yao}.

\textbf{Matching} pairs each treated sample with one or more control samples with similar propensity scores (or covariates) \citep{Peter}. Counterfactual outcomes are then estimated from the matched units:
\[
\begin{split}
	\hat{Y}_i(c) &= 
	\begin{cases}
		Y_i & \text{if } t_i = 0 \\
		\frac{1}{|\mathcal{M}(i)|} \sum_{m \in \mathcal{M}(i)} Y_m & \text{if } t_i = 1
	\end{cases}, \\
	\hat{Y}_i(t) &= 
	\begin{cases}
		\frac{1}{|\mathcal{M}(i)|} \sum_{m \in \mathcal{M}(i)} Y_m & \text{if } t_i = 0 \\
		Y_i & \text{if } t_i = 1
	\end{cases}.
\end{split}
\]

where $\hat{Y}_{i}(c)$ and $\hat{Y}_{i}(t)$ denote the estimated outcomes under control and treatment, respectively, and ${M}_{(i)}$ represents the set of matched or neighboring samples from the opposite treatment group.

\textbf{Overlap weights} \citep{Li} offer another reweighting approach where each unit receives a weight proportional to its probability of receiving the opposite treatment. These weights are bounded in \([0, 1]\), reducing the influence of extreme values and improving estimation stability in regions of covariate overlap.


\section{Literature Review}

Propensity score methods have become central in the literature for addressing selection bias in observational studies. They aim to replicate the balance achieved through randomization by adjusting for covariate differences between treatment groups.

The seminal work on Inverse Propensity Weighting (IPW) \citep{PRrose, IPW} introduced a framework for estimating treatment effects by reweighting observations based on their propensity scores. However, IPW can suffer from instability due to the presence of extreme weights when the estimated propensity scores have high variance.

To address this, researchers have proposed methods such as weight trimming \citep{Lee}, which discards units with excessively large or small propensity scores, and doubly robust estimation \citep{Robins, funk2011doubly}, which combines propensity score modeling with outcome regression, ensuring consistency if either model is correctly specified.

Covariate Balancing Propensity Score (CBPS) \citep{Imai, Christian} redefines propensity score estimation as an optimization problem aimed directly at achieving balance, rather than maximizing likelihood. This innovation improved empirical performance by ensuring covariates are better balanced across groups. Although CBPS reduces sensitivity to model misspecification, it still inherits limitations of inverse propensity weighting, including susceptibility to extreme weights when covariate overlap is limited.

Stratification and matching techniques have also been extensively studied. Stratification divides data based on estimated scores into strata that are assumed to be balanced internally \citep{Imbens, Yao}, while matching methods \citep{Peter} identify comparable units across treatment groups to estimate counterfactuals. These approaches are conceptually intuitive but can struggle with scalability and sensitivity to the choice of distance metric.

Recently, overlap weighting \citep{Li} has been introduced to mitigate the instability of IPW. By weighting units inversely to their treatment assignment probability, this method emphasizes the region of common support between groups, leading to more stable and efficient estimates. However, its effectiveness is contingent on sufficient overlap in the covariate distributions between treatment groups—a condition that may not hold in all observational datasets.

Despite these advances, challenges remain. Many methods break down in settings with poor overlap, high-dimensional covariates, or misspecified models. These limitations motivate our proposed method, which aims to preserve the strengths of reweighting strategies while addressing their known weaknesses.

\section{Deconfounding Factor-based Weighting (DFW)}
This section is organized into four parts. We begin by outlining the problem setup and underlying assumptions. We then provide an intuitive overview of the proposed Deconfounding Factor Weighting (DFW) method, followed by a detailed technical formulation. Finally, we discuss key theoretical properties of the weighting scheme.

\subsection{Formalization and Assumptions}

In practice, we estimate treatment effects using observational data. An observational datasets $\mathcal{D} = \{ \mathbf{x}, t, y\}_{i=1}^N$ consists of pre-treatment variables $\mathbf{x} \in \mathcal{X} \subseteq \mathbb{R}^K$, treatments $t \in \mathcal{T}$ (e.g., {0: medication, 1: surgery}), and observed outcomes $y \in \mathcal{Y}$ (e.g., recovery time). However, only the factual outcomes $y^{t}$ are observed, while the counterfactual outcomes $y^{\neg t}$ remain unobserved \citep{Negar,Khan2024OnTE}. 
Selection bias is present when treatment assignment depends on $\mathbf{x}$, violating the randomized controlled trial assumption $P(\mathcal{T} \mid \mathcal{X}) = P(\mathcal{T})$. Additional details on the estimation of various treatment effect metrics are provided in Section~\ref{EV}. Treatment effect estimation from observational data commonly relies on the following key assumptions \citep{PRrose,Rubin,Imbens}.


\begin{itemize}
	\item \textbf{Stable Unit Treatment Value:} The treatment assigned to one unit does not influence the potential outcomes of any other unit (i.e. patient).
	\item \textbf{Unconfoundedness:} There are no unmeasured confounders; all factors affecting both the treatment $\mathcal{T}$ and the outcome $\mathcal{Y}$ have been observed, formally, $\mathcal{Y} \perp\!\!\!\perp \mathcal{T} \mid \mathcal{X} $.
	
	\item \textbf{Overlap:} The probability of receiving any treatment given covariates $\mathbf{x}$ is strictly positive for all treatments. Formally, $P({t} \mid \mathbf{x}) > 0  \, \forall t \in \mathcal{T}, \forall \mathbf{x} \in \mathcal{X} $.
	
	
\end{itemize}

\subsection{Intuition}
DFW is a propensity score-based weighting scheme designed to address selection bias introduced by confounding variables in observational studies. As the name \emph{Deconfounding Factor Weighting} (DFW) suggests, the method explicitly aims to reduce the confounding impact of such variables using weights. The core idea is to weight each sample by its deconfounding factor, defined as 
\[
1 - P(\mathcal{T} = t \mid \mathcal{X} = \mathbf{x}).
\]
Intuitively, DFW assigns higher weights to samples that are less confounded (i.e., those with a lower probability of receiving their observed treatment) and lower weights to samples that are more confounded. This approach effectively rebalances the data, creating a pseudopopulation that mimics the covariate distribution of a randomized controlled trial (RCT), enabling fair and reliable causal comparisons.

To illustrate, consider Figure \ref{fig:3}, which depicts two treatment groups ($t=0$ and $t=1$) with an unbalanced distribution of age—a known confounding factor. Without adjustment, this imbalance can lead to biased and unreliable causal estimates. DFW addresses this by assigning weights as follows:

\begin{itemize}
	\item In the $t=0$ group, elderly patients (red) (with a 75\% probability of receiving treatment 
	$t=0$) receive a weight of 0.25, while younger patients (with a 25\% probability) receive a weight of 0.75.
	
	\item Conversely, in the $t=1$ group, younger patients (green) (with a 75\% probability of receiving treatment 
	$t=1$) receive a weight of 0.25, while elderly patients (with a 25\% probability) receive a weight of 0.75.
\end{itemize}

As shown in Figure \ref{fig:3}, DFW effectively balances the covariate distribution across treatment groups, creating a pseudopopulation that resembles an RCT. This ensures that the data is suitable for unbiased causal inference.

\subsection{Details of DFW}
In our proposed weighting scheme, each sample is assigned a weight defined as follows:
\begin{equation}
	\begin{split}
		w_{i}= 1- {P}(\mathcal{T}=t_i \mid \mathcal{X}=\mathbf{x}_i) 
	\end{split}
\end{equation}

Here, 
${P}$
denotes the probability of the observed treatment 
$t_i$ given the covariates $\mathbf{x}_i$, estimated using a classifier $f_{cla}$ (primarily trained using logistic regression; further details are provided in Section \ref{sec:results}). Algorithm \ref{alg:algo1} outlines the computation of our scheme's weights, which are then employed for treatment effect estimation via weighted regression $f_{reg}$.
\begin{algorithm}[ht] 
	\caption{Deconfounding Factor-based Weighting (DFW)}
	\label{alg:algo1}
	\textbf{Input}:{ $\mathcal{D}= \{x_1, t_1, y_1\},...,\{x_n, t_n, y_n\}$}\\
	\textbf{Output: }$\mathbf{w} \in \mathbb{R}^n$, $\mathcal{L} \in \mathbb{R}$
	\begin{algorithmic}[1] 
		\STATE train $f_{cla}(\mathbf{x}),f_{reg}(\mathbf{x}) $
		\STATE $\mathbf{w}_{i}\gets1- f_{cla}(\mathbf{x}_i) \quad \forall i$
		\STATE $\mathcal{L}\gets \frac{\sum_{1}^{n} \mathbf{w}_i.l(f_{reg}(\mathbf{x}_i),y)}{\sum_{1}^{n} \mathbf{w}_i} $
		\STATE \textbf{return} $\mathbf{w}, \mathcal{L}$
	\end{algorithmic}
\end{algorithm}

\subsection{Properties}

This subsection highlights two key properties of our Deconfounded Factor Weighting (DFW) scheme: boundedness and low variance. These properties ensure robustness and stability in treatment effect estimation.

\subsubsection{Boundedness:}
Unlike Inverse Probability Weighting (IPW), the weights in our DFW scheme are inherently bounded between 0 and 1. This is achieved by constructing weights based on the deconfounding factor $1- {P}(\mathcal{T}=t \mid \mathcal{X}=\mathbf{x}) $, which naturally restricts the weights to the interval ($0,1$). Formally, the weights satisfy:
\begin{equation}
	0 < w_{i} < 1, 
\end{equation}
This boundedness mitigates the risk of extreme weights, which can destabilize treatment effect estimates in traditional weighting methods.

\subsubsection{Low Variance:} \label{sec:proof}
DFW weights exhibit low variance in weights compared to IPW, even when the characteristics of subjects under study vary significantly. This property ensures that the weights remain stable and do not exhibit drastic fluctuations, leading to better covariate balance and more reliable treatment effect estimation.

Let \( 0 < u < 1 \) be a random variable representing the distribution of propensity scores, sampled from \( p(t \mid \mathcal{X}) \). We are interested in understanding why the variance of the inverse propensity score weights, \( 1/u \), tends to be much larger than the variance of the propensity scores themselves. While our comparison is intuitively between \( 1/u \) and \( 1 - u \), note that \( \mathrm{Var}(1 - u) = \mathrm{Var}(u) \), since variance is invariant under constant shifts. Thus, it is equivalent to comparing the variability of \( 1/u \) with that of \( u \).

To build this intuition, consider the function \( f(u) = 1/u \), which is convex on \( (0, \infty) \). When we transform a variable through a convex function, the variability can become amplified—especially near the boundaries of the domain. Since \( u \) takes values in \( (0,1) \), the function \( 1/u \) grows rapidly as \( u \) approaches zero, leading to large fluctuations in the transformed variable.

We can make this idea more concrete using a second-order Taylor approximation. The variance of a function of a random variable can be approximated as:
\[
\mathrm{Var}(f(u)) \approx [f'(\mathbb{E}[u])]^2 \, \mathrm{Var}(u),
\]
where \( f' \) is the first derivative of \( f \). For \( f(u) = 1/u \), we compute:
\[
f'(u) = -\frac{1}{u^2}, \quad \text{so} \quad [f'(\mathbb{E}[u])]^2 = \frac{1}{(\mathbb{E}[u])^4}.
\]
Since \( \mathbb{E}[u] < 1 \) in typical applications, this amplification factor \( 1/(\mathbb{E}[u])^4 \) is greater than 1. Hence, the variance of \( 1/u \) can be much larger than that of \( u \), especially when small values of \( u \) are likely. This helps explain why inverse propensity weighting can be unstable in practice: low-probability treatment assignments lead to extreme weights and inflated variance.

Empirical evidence supporting this property is presented in Figure \ref{fig:1}.

\section{Experiments} \label{exp}
In this section, we present synthetic and two real-world benchmark datasets used in our experiments and describe the various evaluation criteria employed to assess both covariate balance and treatment effect estimation.

\subsection{Datasets}
In the experiments, our weighing scheme has been evaluated on the following benchmark and synthetic datasets:

\subsubsection{Infant Health and Development Program (IHDP)}
The IHDP dataset is a binary treatment dataset derived from the Infant Health and Development Program study \citep{Gunn}. In \citep{Hill}, selection bias was introduced into the original randomized controlled trial data to convert it into an observational dataset. The dataset comprises 747 instances—139 in the treated group and 608 in the control group—and includes 25 covariates capturing various aspects of the child and mother (e.g., birth weight, neonatal health index, mother's age, and drug status). The primary objective is to evaluate the effect of specialist home visits on the cognitive development of children.

\subsubsection{Jobs}
The Jobs dataset originates from the Lalonde experiment \citep{Hill,lalonde,jobs} and is an observational dataset designed to assess the impact of job training on earnings. It consists of 614 instances, with 185 individuals receiving job training (treated) and 429 in the control group. The dataset includes eight pre-treatment covariates: age, education (\texttt{educ}), race (\texttt{black} and \texttt{hisp}), marital status (\texttt{married}), absence of a degree (\texttt{nodegr}), and prior earnings in 1974 and 1975 (\texttt{re74} and \texttt{re75}). The outcome variable is the earnings in 1978, enabling a thorough evaluation of treatment effectiveness.

\subsubsection{Synthetic datasets}
We generate a synthetic dataset with \( n = 1500 \) samples and \( 6 \) features \( X \sim \mathcal{N}(0, 1) \). The treatment assignment mechanism is defined using a logistic model with selection bias:

\[
P(\mathcal{T}=1 \mid \mathcal{X}) = \sigma(\mathcal{X} \cdot w + \epsilon),
\]

where \( \sigma(\cdot) \) is the sigmoid function, \( w = [w_1, w_2, w_3, w_4, w_5, w_6] \) are the weights inducing selection bias (to vary selection bias, we adjust the weights $w_1,w_4,w_6$), and \( \epsilon \sim \mathcal{N}(0, 0.08) \) adds noise. The treatment \( T \) is sampled from a Bernoulli distribution with probability \( P(\mathcal{T}=1 \mid \mathcal{X} ) \).

The potential outcomes are generated as:
\begin{equation}
	\begin{split}
		&y^0 =  w_1 x_1 + w_2 x_2 - w_3 x_3 + w_4x_4 -w_5 x_5 + w_6 x_6 + \eta_0 \\
		&y^1 = y^0 + c
	\end{split}
\end{equation}

where \( y^0 \) and \( y^1 \) are the potential outcomes under control and treatment, respectively, $\eta_0 \sim \mathcal{N}(0, 0.1)$ represents noise and $c$ is a constant.
The factual outcome \( y_{\text{factual}} \) is observed as:

\[
y_{\text{factual}} = \mathcal{T} \cdot y^1 + (1 - \mathcal{T}) \cdot y^0,
\]

and the counterfactual outcome \( y_{\text{counterfactual}} \) is:

\[
y_{\text{counterfactual}} = (1 - \mathcal{T}) \cdot y^1 + \mathcal{T} \cdot y^0.
\]

	\textbf{Non-linear Data Generation with Heterogeneous Treatment Effect}
	
	We also generate a synthetic dataset with non-linear treatment assignment and outcome mechanisms to evaluate the performance of treatment effect estimation methods under realistic confounding structures and heterogeneous effects.
	
	Each unit \( i \in \{1, \dots, N\} \) is associated with a covariate vector \( \mathbf{x}_i = (x_{i1}, x_{i2}, x_{i3}, x_{i4}, x_{i5}, x_{i6}) \in \mathbb{R}^6 \), where each feature is drawn independently from a standard normal distribution:
	\[
	\mathbf{x}_i \sim \mathcal{N}(0, I_6).
	\]

	The binary treatment assignment \( t_i \in \{0,1\} \) is drawn from a Bernoulli distribution with a non-linear propensity score:
	\[
	e(\mathbf{x}_i) = P(t_i = 1 \mid \mathbf{x}_i) = \sigma\left( \alpha \cdot \tanh(x_{i1}) + \beta \cdot x_{i2}^2 - \gamma \cdot x_{i3} + \varepsilon_i \right),
	\]
	where \( \sigma(z) = \frac{1}{1 + \exp(-z)} \) is the sigmoid function, \( \varepsilon_i \sim \mathcal{N}(0, \sigma^2) \) adds mild noise, and we use \( \alpha = 3.0 \), \( \beta = 1.0 \), \( \gamma = 0.5 \), and \( \sigma = 0.05 \) in our experiments. The treatment assignment is then sampled as:
	\[
	t_i \sim \text{Bernoulli}(e(\mathbf{x}_i)).
	\]

	The potential outcome under control is defined as:
	\[
	y^0_i = 1.5 \cdot x_{i1} + \sin(x_{i2}) - 0.8 \cdot x_{i3} + 0.5 \cdot x_{i4} + \eta_i,
	\]
	where \( \eta_i \sim \mathcal{N}(0, 0.1^2) \) is independent noise.
	
	We introduce heterogeneous treatment effects by defining:
	\[
	\tau_i = 2 + 0.5 \cdot x_{i5},
	\]
	and the potential outcome under treatment is then:
	\[
	y^1_i = y^0_i + \tau_i.
	\]

	The observed (factual) and counterfactual outcomes are defined as:
	\[
	y_i^{\text{factual}} = t_i \cdot y^1_i + (1 - t_i) \cdot y^0_i, \quad
	y_i^{\text{counterfactual}} = (1 - t_i) \cdot y^1_i + t_i \cdot y^0_i.
	\]
	
	This setup enables robust evaluation of causal inference methods in the presence of non-linear selection bias and covariate-dependent heterogeneous treatment effects.

\subsection{Evaluation Criteria} \label{EV}
This subsection outlines three statistical criteria to evaluate the balance of weighted covariate distributions between treated and control groups. Standardized mean difference (SMD) values close to 0 (typically $<$ 0.1) indicate minimal differences in covariate means. The Kolmogorov–Smirnov (KS) test assesses distributional equality, with smaller statistics suggesting better balance. Visual inspection of empirical cumulative distribution function (ECDF) plots provides intuitive confirmation, where overlapping curves indicate well-aligned distributions. Finally, the average treatment effect (ATE) and individual treatment effect (ITE) are evaluated to ensure unbiased causal estimates.

\subsubsection{Standardized Mean Difference (SMD)}

The standardized mean difference (SMD) is a widely used metric for quantifying differences in covariate distributions between treatment groups in observational studies \citep{Peter,SMD}. Expressed in units of standard deviation, the SMD captures the difference in group means relative to the pooled variability. For a given variable, the SMD is defined as:
\[
d = \frac{100 \times (\bar{x}_T - \bar{x}_C)}{\sqrt{\frac{s_T^2 + s_C^2}{2}}},
\]
where \(\bar{x}_T\) and \(\bar{x}_C\) are the sample means in the treated and control groups, respectively, and \(s_T^2\) and \(s_C^2\) denote the corresponding sample variances. In the context of weighted analyses, the weighted mean is computed as
\[
\bar{x}_{\text{weight}} = \frac{\sum_i w_i x_i}{\sum_i w_i},
\]
and the weighted variance is given by
\[
s_{\text{weight}}^2 = \frac{\sum_i w_i}{\left(\sum_i w_i\right)^2 - \sum_i w_i^2} \sum_i w_i \left(x_i - \bar{x}_{\text{weight}}\right)^2.
\]

\subsubsection{Kolmogorov–Smirnov (K-S) Test}
The Kolmogorov–Smirnov (K-S) test is a non-parametric method for comparing the distributions of two samples by quantifying the maximum difference between their empirical distribution functions (EDFs)\citep{KS}. For two samples, the K-S statistic is defined as:
\[
D_{n,m} = \sup_{x} \left| F_{1,n}(x) - F_{2,m}(x) \right|,
\]
where \(F_{1,n}(x)\) and \(F_{2,m}(x)\) are the EDFs for the treated and control groups, comprising \(n\) and \(m\) samples, respectively, and \(\sup_{x}\) denotes the supremum over all \(x\). For further details, please refer to \citep{Joh}.

\subsubsection{Empirical Cumulative Distribution Function (ECDF) Plot}

The ECDF plot is a powerful visualization tool for comparing the distributions of continuous variables. It is based on the empirical cumulative distribution function, a step function that increases by $\frac{1}{N}$ at each observation, thereby representing the observed values and their corresponding percentiles\citep{KS}. Formally, the ECDF is defined as:
\begin{equation}
	F_x(x) = \frac{1}{N}\sum_{i=1}^{N} \mathbf{1}\{x_i \leq x\},
\end{equation}
where $\mathbf{1}\{\cdot\}$ denotes the indicator function. For weighted data, the ECDF is modified to incorporate sample weights as follows:
\begin{equation}
	F_x(x) = \frac{1}{\sum_{i=1}^{N}w_i}\sum_{i=1}^{N}w_i\, \mathbf{1}\{x_i \leq x\}.
\end{equation}

\subsubsection{The Average Treatment Effect (ATE)}
ATE is a measure used in causal inference to estimate the average difference in outcomes between a treatment group and a control group. ATE is defined as \citep{imbens2009recent}\citep{JudeaP}:
\begin{equation}
	ATE = E[Y^1 - Y^0],
\end{equation}
Here,$( Y^1 )$	represents the potential outcome if an individual receives the treatment, $( Y^0)$ 
denotes the potential outcome if they do not, and $( E[\cdot] )$ signifies the expectation (average) over the population. In our experiments we are evaluating $\epsilon_{ATE}$ bias as: $|ATE-\hat{ATE}|$.

\subsubsection{Individual Treatment Effect (ITE)}  

	The Individual Treatment Effect (ITE) measures the difference in potential outcomes at the individual level and is defined as:  
	\[
	\mathrm{ite}_{i} = y^{1}_i - y^{0}_i,
	\]  
	where \(y^1_i\) and \(y^0_i\) denote the potential outcomes for individual \(i\) under treatment and control, respectively.
	
	A widely used metric for evaluating ITE estimation quality is the \emph{Precision in Estimation of Heterogeneous Effect} (PEHE)~\citep{Hill}, defined as:
	\begin{equation}
		\mathit{PEHE} = \sqrt{\frac{1}{N}\sum_{i=1}^{N}(\hat{e}_{i} - e_{i})^2}
	\end{equation}
	where \(e_i = y^1_i - y^0_i\) is the true individual treatment effect, and \(\hat{e}_i = \hat{y}^1_i - \hat{y}^0_i\) is the estimated individual treatment effect, computed from the model's predicted outcomes \(\hat{y}^1_i\) and \(\hat{y}^0_i\) under treatment and control, respectively.
	
	This evaluation metric can only be computed in settings where both potential outcomes are available, such as in semi-synthetic datasets like IHDP, which simulate the counterfactual outcomes for each unit.

\subsubsection{Coefficient of Variation}

The Coefficient of Variation (CV) quantifies relative variability and is defined as:
	\begin{equation}
		CV = \frac{\sigma}{\mu},
	\end{equation}
	where $\sigma$ is the standard deviation and $\mu$ is the mean.
	
	In propensity score weighting, a high CV indicates unstable weights. This is a common issue in Inverse Probability Weighting (IPW), where individuals with very low treatment probabilities receive extremely large weights. Such unbounded weights inflate variance (see Fig.~\ref{fig:1}) and lead to poor covariate balance and inaccurate treatment effect estimates \citep{Cole2008ConstructingIP}.
\section{Results}\label{sec:results}

In this section, we empirically evaluate our approach against baseline methods—Inverse Probability Weighting (IPW), Covariate Balancing Propensity Score (CBPS) and Overlap weighting—across multiple datasets using the previously defined evaluation criteria. We first present results on two real-world benchmark datasets, followed by an analysis on synthetically generated datasets.

For our experiments, we leverage the scikit-learn library for model implementation. We train a logistic regression model ($f_{cla}$) to estimate propensity scores, while Ridge regression and a Support Vector Machine (SVM) regressor ($f_{reg}$) are used for treatment effect estimation in linear and non-linear settings, respectively. To ensure robustness, results are averaged over the first 30 realizations of the IHDP and Jobs datasets, and over 30 independent train/test splits (80/20 ratio) for synthetic datasets.

\subsection{ANALYSIS OF BENCHMARK DATASETS: } This section presents a comprehensive analysis of the experimental results on benchmark datasets i.e. IHDP and Jobs.

\textbf{SMD:} Figure~\ref{fig:4} compares our approach with baseline methods using standardized mean difference (SMD) as the evaluation metric. SMD (lower is better) is computed between treated and control groups across six continuous covariates in the IHDP dataset.

The left panel of Figure~\ref{fig:4} shows that Inverse Probability Weighting (IPW) reduces SMD below the 10 threshold for some variables, improving upon the unweighted case. The middle panel illustrates the performance of Covariate Balancing Propensity Score (CBPS), which struggles to balance certain covariates. The Overlap weighting method, shown alongside, performs well on IHDP, with only two covariates exceeding the threshold, indicating effective covariate balance. Finally, the right panel demonstrates that our DFW approach achieves the strongest covariate balance---SMD values are consistently below 10, with most under 5---outperforming all baselines.

A similar trend holds in Figure~\ref{fig:5}, which presents results on the Jobs dataset. While IPW and CBPS fail to reduce SMD to acceptable levels, both DFW and Overlap weighting provide substantial improvements. In particular, Overlap is highly competitive and comparable to DFW, with both methods ensuring that most covariates are below the threshold or very close to it---highlighting their effectiveness in achieving covariate balance under practical, real-world conditions.

\textbf{ECDF Plot: }
Figure~\ref{fig:6} presents empirical cumulative distribution function (ECDF) plots for the birth order covariate in the IHDP dataset under different weighting schemes. The first three panels illustrate that the treated (blue) and control (orange) groups exhibit significantly different distributions when weighted using IPW, CBPS, and Overlap, respectively. In contrast, the last panel—based on our DFW weighting scheme—shows near-identical distributions between treated and control groups, indicating superior covariate balance.

Similarly, Figure~\ref{fig:7} compares ECDF plots across different methods for the Jobs dataset. The first three panels reveal notable discrepancies in treated and control distributions when using IPW, CBPS, and Overlap weights. However, the final panel demonstrates that after applying DFW, the distributions align almost perfectly. While Overlap performs better than IPW and CBPS, its results remain slightly inferior to DFW.

\textbf{K-S Test: } Tables~\ref{tab:2} and \ref{tab:3} report the Kolmogorov--Smirnov (K-S) statistics with 95\% confidence intervals for selected features from the IHDP and Jobs datasets, respectively. These statistics quantify the maximum discrepancy between the empirical distributions of treated and control groups, with lower values indicating better covariate balance.

On the \textbf{IHDP dataset} (Table~\ref{tab:2}), DFW achieves the lowest K-S values on three out of six features (\texttt{birth.w}, \texttt{preterm}, and \texttt{birth.o}), and performs competitively on the remaining features. Overlap weighting achieves the best performance on \texttt{nnhealth} and \texttt{momage}, though it exhibits wider confidence intervals on other features. IPW and CBPS show greater variability, with generally higher K-S values and broader uncertainty ranges.

On the \textbf{Jobs dataset} (Table~\ref{tab:3}), DFW again demonstrates competitive performance, achieving the best covariate balance on \texttt{x1} and \texttt{x5}, and performing nearly on par with Overlap on \texttt{x2} and \texttt{x4}. For \texttt{x3}, CBPS attains the lowest point estimate, but the wide confidence interval indicates less stable behavior. 

Across both datasets, DFW consistently provides strong balance performance with narrower or comparable confidence intervals, indicating both accuracy and robustness. These results highlight DFW's ability to achieve reliable covariate balance across diverse datasets and variable types, often outperforming or matching established methods such as IPW, CBPS, and Overlap weighting.

\textbf{Treatment Effect Estimation: } Table~\ref{tab:4} presents the performance of four weighting methods---IPW, CBPS, Overlap, and DFW---on treatment effect estimation, evaluated using the mean absolute bias in average treatment effect ($\epsilon_{ATE}$) and Precision in Estimation of Heterogeneous Effect (PEHE). Results are reported for both linear and non-linear outcome models on the IHDP and Jobs datasets.

Under the linear outcome model, DFW achieves the lowest $\epsilon_{ATE}$ on both IHDP (0.34) and Jobs (0.09), outperforming all baselines by a clear margin. For PEHE on IHDP, DFW also reports the smallest error (1.32), significantly lower than CBPS (3.80) and Overlap (3.78), demonstrating its accuracy in estimating individual-level effects.

In the non-linear setting, DFW maintains its strong performance, achieving the lowest $\epsilon_{ATE}$ on IHDP (0.33, tied with IPW) and matching the best result on Jobs (0.11), alongside CBPS and Overlap. For PEHE, DFW again performs best (1.29) on IHDP, showing improved robustness even under more complex outcome specifications.

Overall, DFW consistently delivers accurate ATE and PEHE estimates across datasets and outcome models, highlighting its effectiveness in mitigating selection bias and improving treatment effect estimation.

\subsection{ANALYSIS OF SYNTHETIC DATASETS: } This section evaluates the performance of the proposed method on synthetic datasets.

\textbf{SMD: } Figure~\ref{fig:2} presents the Standardized Mean Difference (SMD) analysis across two synthetic datasets, where the upper panel corresponds to a dataset with low selection bias, and the lower panel represents a high selection bias setting. To visually illustrate the degree of selection bias, we show the distribution of covariate 4 in the figure.

From Figure~\ref{fig:2}, it is evident that the performance of IPW, CBPS, and Overlap deteriorates as selection bias increases, failing to consistently reduce SMD below the desired threshold. In contrast, our proposed DFW method remains robust across both settings, effectively controlling SMD values within an acceptable range. This demonstrates that DFW maintains superior covariate balance even under strong selection bias, outperforming the baseline methods.

\textbf{ECDF plot: } Figure~\ref{fig:8} presents the ECDF plot of covariate 4 from the high selection bias synthetic dataset (selected due to space constraints). The results further reinforce our findings: IPW, CBPS, and Overlap struggle to align the distributions between treatment groups, whereas DFW effectively balances them, resulting in closely overlapping empirical distributions.

\textbf{Treatment Effect Estimation: } Table~\ref{tab:1} presents the mean absolute bias in average treatment effect ($\epsilon_{ATE}$) across synthetic datasets with varying levels of selection bias. Lower values indicate more accurate estimation. In the low-bias setting, DFW outperforms all other methods, achieving the lowest error (5.54), while Overlap and CBPS yield similar results (7.58 and 7.52, respectively), both outperforming IPW (12). 

As selection bias increases, IPW's performance deteriorates dramatically, with extreme errors in the moderately and highly biased datasets (3.12e$^{12}$ and 6.98e$^{13}$), rendering it ineffective. This sharp decline underscores IPW's inability to achieve covariate balance when units exhibit diverse characteristics and some have low propensity scores—i.e., when the positivity or overlap assumption is weakly satisfied. CBPS remains more stable (errors of 9.73 and 10.31), but consistently lags behind DFW. Overlap achieves slightly better accuracy than CBPS in moderate and high bias settings (9.23 and 9.34), though DFW consistently yields the lowest $\epsilon_{ATE}$ (7.19 and 7.27), highlighting its robustness to selection bias and superior reliability in treatment effect estimation.

For non-linear models, a similar trend is observed, though the overall errors are more controlled across methods. DFW continues to outperform others, achieving the lowest $\epsilon_{ATE}$ in all bias settings (4.85, 6.45, and 6.40). Overlap performs better than CBPS in moderate and high bias regimes (6.65 and 6.88 vs.\ 8.63 and 13.05), and remains close to DFW in low bias (5.50). While IPW does not suffer from the extreme numerical instability seen in the linear model, it still yields larger errors (5.28–6.86) compared to Overlap and DFW.

\textbf{K-S test: }
Table~\ref{tab:5} presents the Kolmogorov–Smirnov (K-S) statistics with 95\% confidence intervals for six covariates across synthetic datasets under low, moderate, and high levels of selection bias. The results compare the performance of four reweighting methods: IPW, CBPS, Overlap, and DFW. Lower K-S values indicate better covariate balance.

In the \textit{low bias setting}, DFW consistently achieves the best covariate balance on five out of six features. For instance, on feature \texttt{x5}, DFW attains a K-S statistic of \textbf{0.191} {\tiny[0.095, 0.290]}, compared to 0.313 for IPW and 0.223 for CBPS. Similarly, on \texttt{x3} and \texttt{x4}, DFW’s values of \textbf{0.271} and \textbf{0.231}, respectively, are the lowest, with Overlap and CBPS slightly trailing behind. Only on \texttt{x0} does CBPS slightly outperform DFW, with 0.274 {\tiny[0.185, 0.381]} vs.\ 0.280 {\tiny[0.169, 0.357]}, although their confidence intervals largely overlap. These results suggest that all methods are relatively effective under mild bias, but DFW provides the most consistently low dispersion.

As selection bias increases to a \textit{moderate level}, DFW maintains its lead across all six covariates. For example, on feature \texttt{x1}, DFW records a K-S value of \textbf{0.203} {\tiny[0.115, 0.353]}, clearly outperforming CBPS (0.292) and IPW (0.454). The difference is more substantial on features such as \texttt{x3} and \texttt{x5}, where IPW values rise to 0.566 and 0.538, respectively, while DFW remains comparatively low at \textbf{0.363} and \textbf{0.382}. These trends reflect DFW's robustness in maintaining balance even as the propensity score distribution becomes more skewed.

In the \textit{high bias scenario}, DFW again yields the lowest K-S statistic on every feature. The contrast is particularly striking for \texttt{x4}, where DFW achieves \textbf{0.155} {\tiny[0.087, 0.246]}, whereas IPW reaches 0.530 {\tiny[0.158, 0.976]}, indicating nearly a threefold increase in imbalance for IPW. On feature \texttt{x2}, DFW’s value of \textbf{0.184} is significantly lower than CBPS (0.338) and IPW (0.555), and comparable to Overlap (0.188). Notably, DFW’s estimates remain stable even as other methods degrade, highlighting its reliability in high-confounding scenarios.

Overall, these results demonstrate that DFW provides superior covariate balance not only in low-bias conditions but especially under moderate and high selection bias. Its performance remains consistent across features and bias levels, suggesting its effectiveness in producing well-balanced samples for reliable treatment effect estimation.

\textbf{Non-linear data settings with heterogeneous treatment effect:} 
Table~\ref{tab:6} presents the mean absolute bias in average treatment effect ($\epsilon_{ATE}$) under non-linear data settings across synthetic datasets with varying levels of selection bias. Lower values indicate better treatment effect estimation. In the \textbf{low bias} setting, both DFW and IPW achieve the lowest error (1.31), substantially outperforming CBPS (2.67) and Overlap (2.72). This suggests that when the selection bias is limited, simple models like IPW can still perform well; however, DFW maintains this performance even as complexity increases.
	
	In the \textbf{moderately biased} setting, DFW again provides the best performance with an error of 1.93, slightly improving over IPW (1.95), and significantly outperforming CBPS (3.34) and Overlap (3.36). This highlights DFW’s robustness in scenarios with increased confounding.
	
	In the \textbf{high bias} setting, IPW breaks down entirely, yielding a large error (5.55e$^{11}$), indicating a failure due to poor overlap. In contrast, DFW maintains a low error (2.24), significantly outperforming both CBPS (3.78) and Overlap (3.66). This result emphasizes DFW’s stability under severe selection bias and its ability to maintain reliable treatment effect estimates where traditional methods fail.

Figure \ref{fig:9} illustrates the effectiveness of our method in separating feature contributions on a synthetic dataset \cite{Khan2024OnTE} with predefined feature roles. As expected, the classifier relies only on instrumental and confounding features for propensity estimation, while the regression model utilizes confounding and adjustment features for outcome prediction with DFW weights. This confirms that our approach correctly disentangles feature influences as intended.

We evaluated our method and IPW across a broad range of propensity scores, spanning from 0.10 to 0.90 in increments of 0.10, considering all possible combinations of six samples. The results indicate that our method consistently yields lower coefficient of variation (CV) in weights, leading to improved covariate balance. As shown in Figure \ref{fig:1}, the CV difference remains negative for almost 75\% of the combinations. This demonstrates that our method maintains lower variability in weights compared to IPW, reinforcing its effectiveness in stabilizing treatment effect estimation.

\subsection{Computational complexity}
	
	We compare the computational complexity of DFW to IPW, CBPS, and Overlap weighting. Both IPW and Overlap weighting involve estimating propensity scores—typically via logistic regression—followed by simple post-processing to compute sample weights. This results in a computational complexity of approximately $\mathcal{O}(n p)$, where $n$ is the number of samples and $p$ is the number of covariates.
	
	DFW similarly employs logistic regression. Since it avoids complex model architectures or nested optimization procedures, its overall complexity remains on par with IPW and Overlap weighting, i.e., $\mathcal{O}(n p)$. Therefore, DFW offers improved robustness without incurring additional computational cost.
	
	In contrast, CBPS formulates a constrained optimization problem that simultaneously balances covariates and models treatment assignment. This involves solving a system of moment conditions via iterative solvers, with complexity increasing with the number of covariates and requiring additional convergence checks. As a result, CBPS is generally more computationally intensive than the other methods, especially in high-dimensional settings.

\subsection{DFW under extreme model misspecification}
	While DFW is designed to improve robustness in settings with selection bias and limited covariate overlap, it is not immune to the effects of extreme model misspecification.
	In particular, DFW relies on estimating the deconfounding factor using a model (e.g., logistic regression). If this model is severely misspecified—for example, if the true treatment assignment mechanism involves complex nonlinear or high-order interactions that are not captured—then the resulting weights may fail to adequately balance confounders across treatment groups. This can lead to residual bias in treatment effect estimates, similar to the behavior observed in other weighting-based methods under extreme misspecification. Exploring model selection and regularization strategies to enhance robustness under misspecification is a valuable direction for future work.

\section{Conclusion}
Causal inference from observational data requires careful handling of selection bias to ensure valid treatment effect estimates. While inverse probability weighting (IPW) is a widely used technique, its reliance on underlying data and model for propensity scores estimation leads to unbounded weights, increased variance, and suboptimal performance. We introduced Deconfounding Factor Weighting (DFW), a novel weighting scheme that reweights samples using the deconfounding factor instead of the inverse propensity score. DFW effectively balances covariates while maintaining bounded, stable weights, reducing variance, and improving the reliability of causal effect estimation. Our empirical evaluation on benchmark datasets confirms that DFW outperforms standard approaches in both covariate balancing and treatment effect estimation. Given its empirical advantages, DFW provides a scalable and effective alternative for causal inference in real-world scenarios.


\section*{Author Contributions}

[Ahmad Saeed Khan] is the primary author and conducted the main research, including formulating the research questions, developing the methodology, implementing the experiments, and writing the manuscript.

[Erik Schaffernicht] provided feedback on the manuscript and offered critical insights during the development and evaluation phases.

[Johannes A. Stork] supervised the project, provided guidance on the research design and analysis, assisted in refining the methodology, supported in writing and contributed to the interpretation of results and manuscript revisions.

\section*{Funding}
This work has been funded by the Industrial Graduate School Collaborative AI \& Robotics, the Swedish Knowledge Foundation Dnr\:20190128, and the Knut and Alice Wallenberg Foundation through Wallenberg AI, Autonomous Systems and Software Program (WASP).

\section*{Data Availability Statement}
The datasets analyzed and generated for this study can be found in the https://github.com/askhanatgithub/DFW

\bibliographystyle{Frontiers-Harvard} 
\bibliography{test}

\begin{table}[!h]
	\centering
	\small
	\addtolength{\tabcolsep}{-4pt}
	\begin{tabular}{lllll} 
		
		\toprule
		Dataset  & \multicolumn{4}{c}{$\epsilon_{ATE}$ Linear model}  \\
		\cmidrule(lr){2-5} 
		& IPW & CBPS & Overlap & DFW  \\
		\midrule
		Low biased  & 12\tiny{(0.00)} & 7.52\tiny{(0.00)} & 7.58 (0.00)& \textbf{5.54}\tiny{(3.95)} \\
		Moderately biased  & 3.12e$^{12}$\tiny{(1.31e$^{12}$)} & 9.73\tiny{(0.00)} &9.23(0.00) & \textbf{7.19}\tiny{(0.00)} \\
		Highly biased  & {6.98e$^{13}$}\tiny{(8.14e$^{12}$)} & 10.31\tiny{(0.00)} & 9.34(0.00)& \textbf{7.27}\tiny{(0.00)} \\
		\midrule
		& \multicolumn{4}{c}{$\epsilon_{ATE}$ Non-linear model}  \\
		\cmidrule(lr){2-5} 
		& IPW & CBPS &Overlap & DFW  \\
		\midrule
		Low biased  & 5.28\tiny{(0.48)} & 5.11\tiny{(0.71)} & 5.50(0.54)& \textbf{4.85}\tiny{(0.49)} \\
		Moderately biased  & 6.80\tiny{(0.40)} & 8.63\tiny{(1.75)} & 6.65(0.45)& \textbf{6.45}\tiny{(0.43)} \\
		Highly biased  & 6.86\tiny{(0.46)} & 13.05\tiny{(3.28)} & 6.88(0.53)& \textbf{6.40}\tiny{(0.48)} \\
		\bottomrule
	\end{tabular}
	\caption{Treatment effect estimation (smaller the better) on synthetic datasets.}
	\label{tab:1}
\end{table}

\begin{table}[!htb]
	\centering
	
	\begin{tabular}{lcccc}
		\toprule
		\textbf{Feature} & \textbf{IPW} & \textbf{CBPS} & \textbf{Overlap} & \textbf{DFW} \\
		\midrule
		birth.w   & 0.231 {\tiny [0.105, 0.419]} & 0.236 {\tiny [0.097, 0.487]} & 0.283 {\tiny [0.113, 0.565]} & \textbf{0.216} {\tiny [0.102, 0.430]} \\
		birth.h   & 0.189 {\tiny [0.076, 0.424]} & 0.175 {\tiny [0.043, 0.380]} & \textbf{0.118} {\tiny [0.019, 0.292]} & 0.156 {\tiny [0.047, 0.317]} \\
		preterm   & 0.269 {\tiny [0.132, 0.452]} & 0.269 {\tiny [0.132, 0.556]} & 0.302 {\tiny [0.129, 0.626]} & \textbf{0.249} {\tiny [0.131, 0.393]} \\
		birth.o   & 0.276 {\tiny [0.133, 0.506]} & 0.277 {\tiny [0.110, 0.603]} & \textbf{0.208} {\tiny [0.104, 0.381]} & 0.236 {\tiny [0.114, 0.436]} \\
		nnhealth  & 0.149 {\tiny [0.005, 0.319]} & 0.125 {\tiny [0.005, 0.415]} & \textbf{0.004} {\tiny [0.000, 0.065]} & 0.125 {\tiny [0.004, 0.298]} \\
		momage    & 0.103 {\tiny [0.006, 0.333]} & 0.057 {\tiny [0.000, 0.149]} & \textbf{0.001} {\tiny [0.000, 0.019]} & 0.072 {\tiny [0.003, 0.204]} \\
		\bottomrule
	\end{tabular}
	\caption{K-S statistics (\tiny 95\% CI\normalsize) for covariate balance on IHDP features after weighting. Lower values indicate better balance.}
	\label{tab:2}
\end{table}

\begin{table}[!htb]
	\centering
	
	\begin{tabular}{lcccc}
		\toprule
		\textbf{Feature} & \textbf{IPW} & \textbf{CBPS} & \textbf{Overlap} & \textbf{DFW} \\
		\midrule
		x1 & 0.500 {\tiny [0.412, 0.672]} & 0.522 {\tiny [0.309, 0.733]} & 0.412 {\tiny [0.296, 0.536]} & \textbf{0.411} {\tiny [0.303, 0.532]} \\
		x2 & 0.391 {\tiny [0.245, 0.537]} & 0.406 {\tiny [0.192, 0.599]} & \textbf{0.363} {\tiny [0.212, 0.529]} & 0.365 {\tiny [0.225, 0.525]} \\
		x3 & 0.520 {\tiny [0.352, 0.665]} & \textbf{0.457} {\tiny [0.098, 0.672]} & 0.483 {\tiny [0.277, 0.661]} & 0.483 {\tiny [0.280, 0.658]} \\
		x4 & \textbf{0.043} {\tiny [0.002, 0.204]} & 0.096 {\tiny [0.002, 0.406]} & 0.054 {\tiny [0.005, 0.129]} & 0.055 {\tiny [0.005, 0.135]} \\
		x5 & {0.594} {\tiny [0.366, 0.712]} & 0.637 {\tiny [0.381, 0.782]} & 0.599 {\tiny [0.517, 0.670]} & \textbf{0.593} {\tiny [0.513, 0.674]} \\
		\bottomrule
	\end{tabular}
	\caption{K-S statistics (\tiny 95\% CI\normalsize) for the first five features in the Jobs dataset after weighting. Lower values indicate better covariate balance.}
	\label{tab:3}
\end{table}

\begin{table}[!htb]
	\centering
	\small
	\addtolength{\tabcolsep}{-4pt}
	\begin{tabular}{lllllllll} 
		\toprule
		\multicolumn{9}{c}{Linear model} \\
		Dataset  & \multicolumn{4}{c}{$\epsilon_{ATE}$} & \multicolumn{4}{c}{PEHE} \\
		\cmidrule(lr){2-5} \cmidrule(lr){6-9}
		& IPW & CBPS &Overlap & DFW  & IPW & CBPS &Overlap& DFW \\
		\midrule
		IHDP  & 0.42\tiny{(0.00} & 0.56\tiny{(0.00)} &0.61\tiny{(0.00)}& \textbf{0.34}\tiny{(0.00)} & 1.50\tiny{(0.01)} & 3.80\tiny{(0.00)} &3.78\tiny{(0.00)}& \textbf{1.32}\tiny{(0.00)} \\
		Jobs  & 160\tiny{(35)} & 0.19\tiny{(0.00)} &0.17\tiny{(0.00)}& \textbf{0.09}\tiny{(0.00)} & NA & NA &NA& NA \\
		\midrule
		\multicolumn{9}{c}{Non-linear model} \\
		& \multicolumn{4}{c}{$\epsilon_{ATE}$} & \multicolumn{4}{c}{PEHE} \\
		\cmidrule(lr){2-5} \cmidrule(lr){6-9}
		& IPW & CBPS &Overlap& DFW  & IPW & CBPS &Overlap& DFW \\
		\midrule
		IHDP  & 0.33\tiny{(0.29)} & 0.37\tiny{(0.29)} &0.38\tiny{(0.29)}& \textbf{0.33}\tiny{(0.29)} & 1.49\tiny{(0.70)} & 2.51\tiny{(3.99)} &2.47\tiny{(3.92)}& \textbf{1.29}\tiny{(0.71)} \\
		Jobs  & 0.17\tiny{(0.11)} & \textbf{0.11}\tiny{(0.07)}& \textbf{0.11}\tiny{(0.07)} & \textbf{0.11}\tiny{(0.08)} & NA & NA &NA& NA \\
		\bottomrule
	\end{tabular}
	\caption{Treatment effect estimation (smaller the better) on IHDP and Jobs datasets.}
	\label{tab:4}
\end{table}

\begin{table}[!htb]
	\centering
	\small
	\begin{tabular}{llcccc}
		\toprule
		\textbf{Bias Level} & \textbf{Feature} & \textbf{IPW} & \textbf{CBPS} & \textbf{Overlap} & \textbf{DFW} \\
		\midrule
		
		\multirow{6}{*}{Low} 
		& x0 & 0.323 {\tiny [0.215, 0.474]} & \textbf{0.274} {\tiny [0.185, 0.381]} & 0.280 {\tiny [0.167, 0.362]} & 0.280 {\tiny [0.169, 0.357]} \\
		& x1 & 0.443 {\tiny [0.241, 0.630]} & 0.389 {\tiny [0.225, 0.457]} & {0.363} {\tiny [0.282, 0.439]} & \textbf{0.361} {\tiny [0.281, 0.438]} \\
		& x2 & 0.366 {\tiny [0.182, 0.674]} & 0.314 {\tiny [0.187, 0.489]} & 0.310 {\tiny [0.221, 0.389]} & \textbf{0.309} {\tiny [0.222, 0.385]} \\
		& x3 & 0.360 {\tiny [0.195, 0.671]} & 0.295 {\tiny [0.206, 0.455]} & 0.273 {\tiny [0.196, 0.371]} & \textbf{0.271} {\tiny [0.194, 0.367]} \\
		& x4 & 0.299 {\tiny [0.114, 0.408]} & 0.238 {\tiny [0.134, 0.328]} & 0.232 {\tiny [0.102, 0.371]} & \textbf{0.231} {\tiny [0.099, 0.367]} \\
		& x5 & 0.313 {\tiny [0.179, 0.524]} & 0.223 {\tiny [0.123, 0.325]} & 0.192 {\tiny [0.094, 0.293]} & \textbf{0.191} {\tiny [0.095, 0.290]} \\
		
		\midrule
		
		\multirow{6}{*}{Moderate} 
		& x0 & 0.526 {\tiny [0.246, 0.911]} & 0.432 {\tiny [0.279, 0.671]} & {0.374} {\tiny [0.241, 0.499]} & \textbf{0.374} {\tiny [0.245, 0.497]} \\
		& x1 & 0.454 {\tiny [0.218, 0.894]} & 0.292 {\tiny [0.140, 0.586]} & 0.204 {\tiny [0.111, 0.356]} & \textbf{0.203} {\tiny [0.115, 0.353]} \\
		& x2 & 0.510 {\tiny [0.198, 0.961]} & 0.310 {\tiny [0.130, 0.669]} & 0.233 {\tiny [0.142, 0.358]} & \textbf{0.232} {\tiny [0.144, 0.353]} \\
		& x3 & 0.566 {\tiny [0.260, 0.975]} & 0.437 {\tiny [0.288, 0.737]} & {0.364} {\tiny [0.233, 0.466]} & \textbf{0.363} {\tiny [0.235, 0.466]} \\
		& x4 & 0.498 {\tiny [0.132, 0.910]} & 0.279 {\tiny [0.113, 0.600]} & {0.164} {\tiny [0.085, 0.319]} & \textbf{0.163} {\tiny [0.087, 0.319]} \\
		& x5 & 0.538 {\tiny [0.279, 0.890]} & 0.404 {\tiny [0.259, 0.661]} & 0.384 {\tiny [0.268, 0.486]} & \textbf{0.382} {\tiny [0.266, 0.480]} \\
		
		\midrule
		
		\multirow{6}{*}{High} 
		& x0 & 0.641 {\tiny [0.307, 0.966]} & 0.480 {\tiny [0.254, 0.771]} & {0.408} {\tiny [0.261, 0.549]} & \textbf{0.407} {\tiny [0.260, 0.542]} \\
		& x1 & 0.468 {\tiny [0.180, 0.940]} & 0.283 {\tiny [0.124, 0.645]} & 0.203 {\tiny [0.094, 0.349]} & \textbf{0.200} {\tiny [0.089, 0.335]} \\
		& x2 & 0.555 {\tiny [0.186, 0.966]} & 0.338 {\tiny [0.121, 0.687]} & 0.188 {\tiny [0.093, 0.311]} & \textbf{0.184} {\tiny [0.094, 0.306]} \\
		& x3 & 0.690 {\tiny [0.464, 0.985]} & 0.509 {\tiny [0.340, 0.799]} & {0.399} {\tiny [0.316, 0.497]} & \textbf{0.398} {\tiny [0.314, 0.496]} \\
		& x4 & 0.530 {\tiny [0.158, 0.976]} & 0.313 {\tiny [0.089, 0.715]} & {0.157} {\tiny [0.088, 0.253]} & \textbf{0.155} {\tiny [0.087, 0.246]} \\
		& x5 & 0.681 {\tiny [0.346, 0.987]} & 0.505 {\tiny [0.313, 0.824]} & {0.428} {\tiny [0.327, 0.547]} & \textbf{0.424} {\tiny [0.323, 0.537]} \\
		
		\bottomrule
	\end{tabular}
	\caption{K-S statistics (\tiny 95\% CI\normalsize) for covariate balance on synthetic datasets under low, moderate, and high bias. Best results per feature are bolded.}
	\label{tab:5}
\end{table}

\begin{table}[!h]
	\centering
	\small
	\addtolength{\tabcolsep}{-4pt}
	\begin{tabular}{lllll} 
		
		\toprule
		\multicolumn{5}{c}{$\epsilon_{ATE}$}\\
		\midrule
		Dataset & IPW & CBPS & Overlap & DFW  \\
		\midrule
		Low biased  & 1.31\tiny{(0.00)} & 2.67\tiny{(0.00)} & 2.72(0.00)& \textbf{1.31}\tiny{(0.00)} \\
		Moderately biased  & 1.95\tiny{(0.0)} & 3.34\tiny{(0.00)} &3.36(0.00) & \textbf{1.93}\tiny{(0.00)} \\
		Highly biased  & {5.55e$^{11}$}\tiny{(2.37e$^{11}$)} & 3.78\tiny{(0.00)} & 3.66(0.00)& \textbf{2.24}\tiny{(0.00)} \\
		
		\bottomrule
	\end{tabular}
	\caption{Treatment effect estimation (smaller the better) on synthetic datasets (non linear data settings).}
	\label{tab:6}
\end{table}

\begin{figure}[!htb]
	\centering
	\setcounter{figure}{0}
	\includegraphics[width=0.45\textwidth]{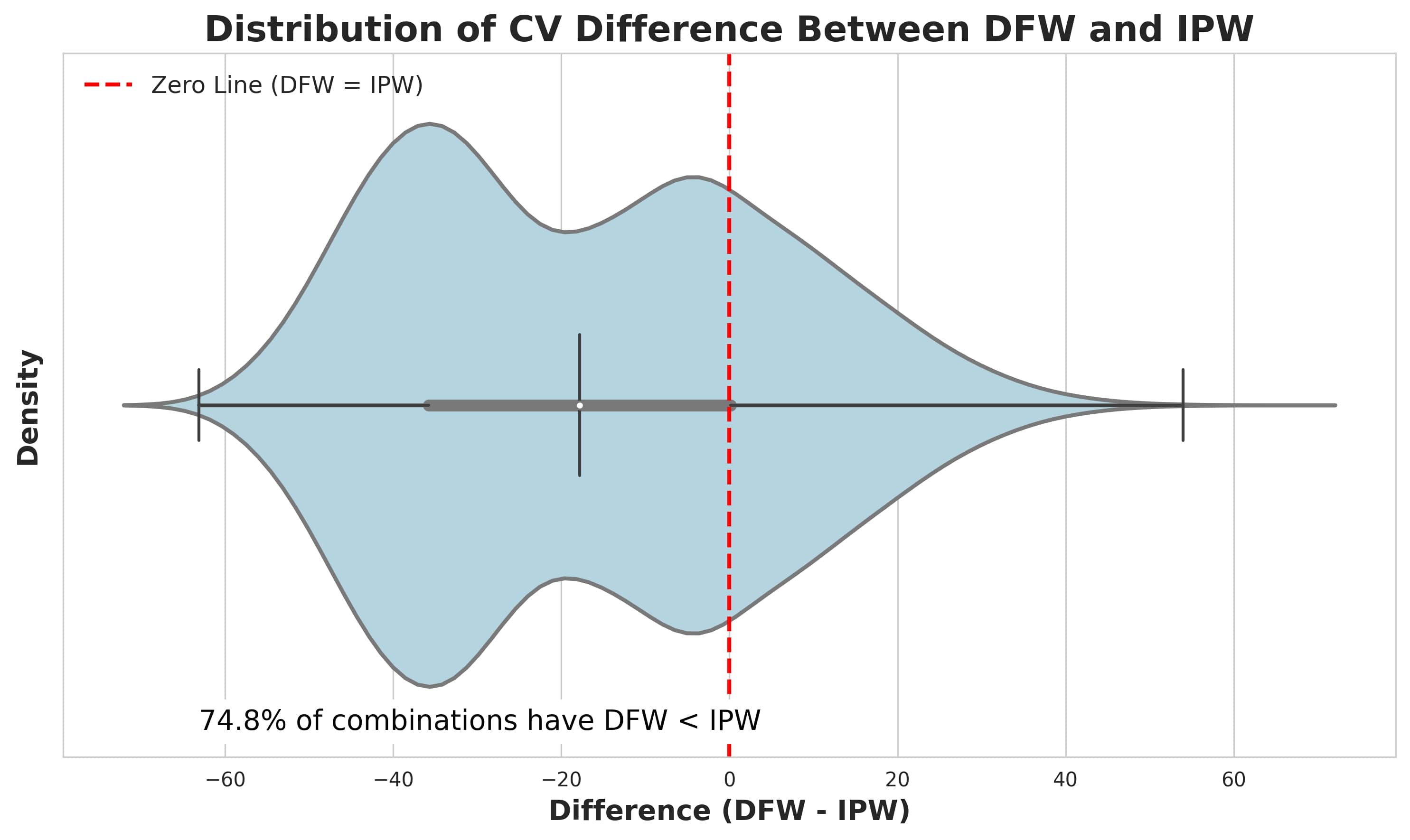}
	\caption{Distribution of CV the differences between the DFW and Inverse Probability Weighting (IPW) calculated across all samples. The violin+box plot shows the density of differences, with a red dashed line indicating where the CVs are equal (DFW = IPW). The annotation indicates that 74.8\% of the combinations have CV of DFW $<$ IPW, highlighting the intrinsic ability of DFW in maintaining low variance in weights.}
	\label{fig:1}
\end{figure}

\begin{figure*}[!htb]
	\centering
	\setcounter{figure}{1}
	\includegraphics[width=0.99\textwidth]{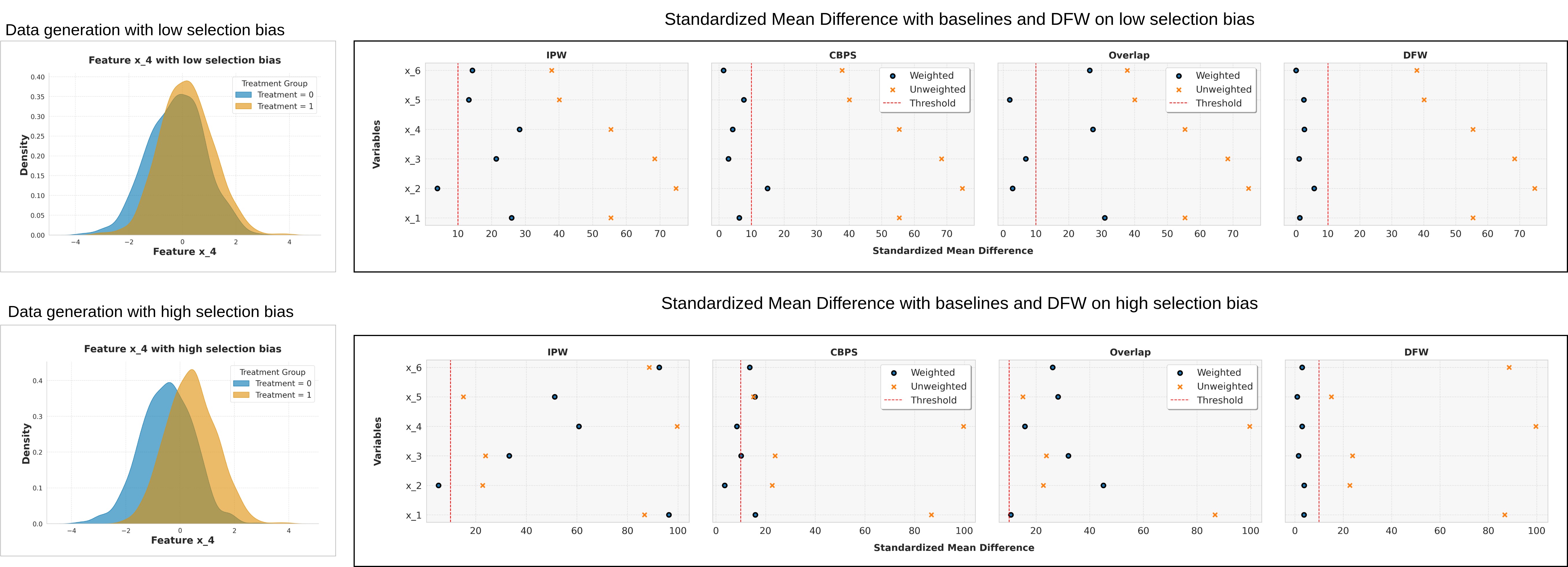}
	\caption{Absolute Standardized Mean Difference (SMD) analysis on two synthetic datasets with varying levels of selection bias. The upper panel corresponds to a dataset with low selection bias, while the lower panel represents a high selection bias setting (visualized using covariate 4). As selection bias increases, the performance of IPW, CBPS, and Overlap weighting deteriorates, failing to consistently reduce SMD below the desired threshold. In contrast, our proposed DFW method remains robust, effectively controlling SMD values across both scenarios.}
	\label{fig:2}
\end{figure*}

\begin{figure}[!htb]
	\centering
	\setcounter{figure}{2}
	\includegraphics[width=0.38\textwidth]{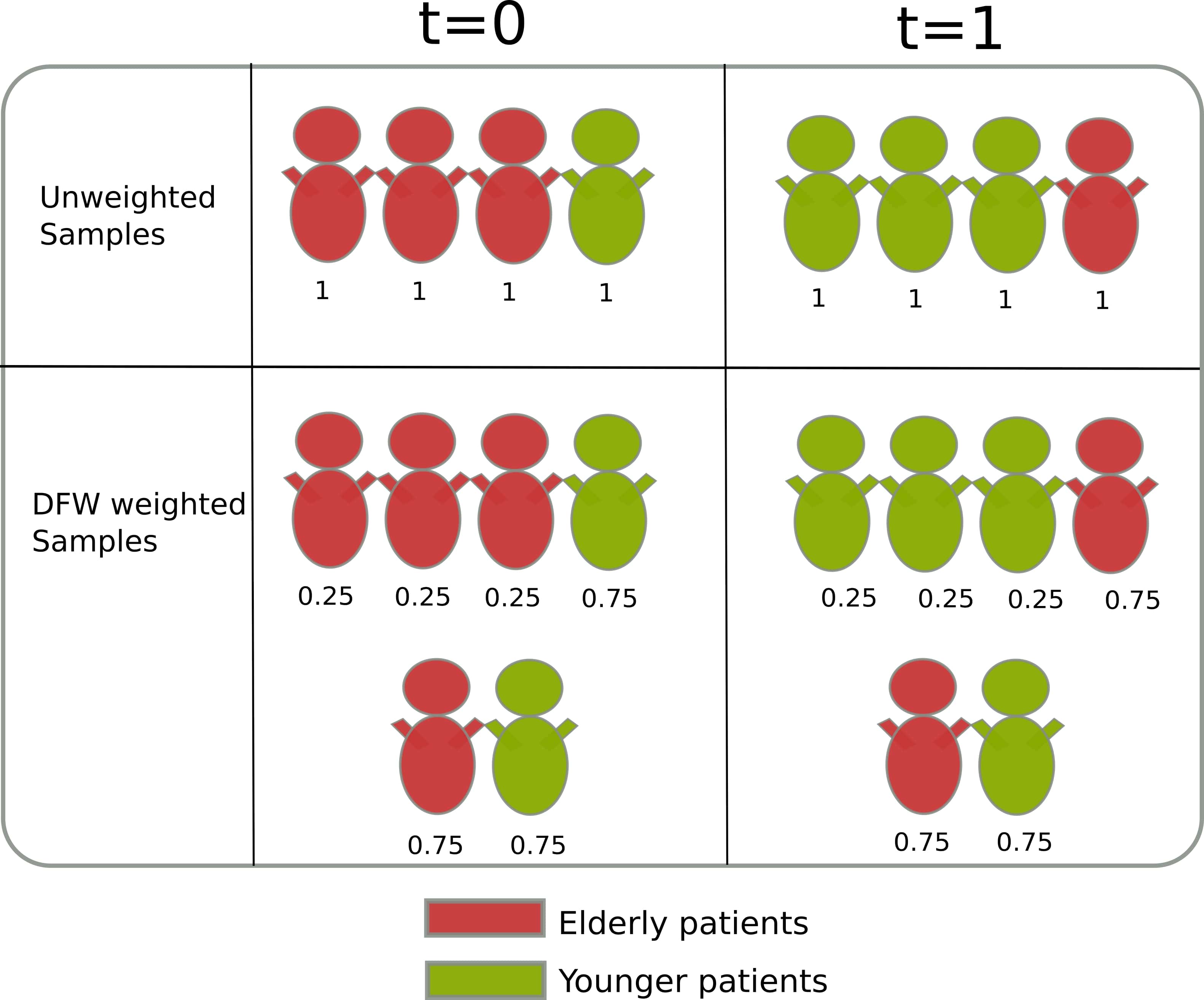}
	\caption{The figure shows two treatment groups ($t=0$ and $t=1$) with an imbalanced age distribution—a key confounder. DFW corrects this by weighting samples: $1$-the probability of receiving the observed treatment. For instance, in the t=0 group, elderly patients (red) (75\% probability) are assigned a weight of 0.25 while younger patients (green) (25\% probability) receive 0.75; the pattern reverses in the t=1 group. This process yields a balanced pseudo-population akin to an RCT, enabling unbiased causal inference by making age distribution equal in both treatment groups.}
	\label{fig:3}
\end{figure}

\begin{figure*}[!htb]
	\centering
	\setcounter{figure}{3}
	\includegraphics[width=1\textwidth]{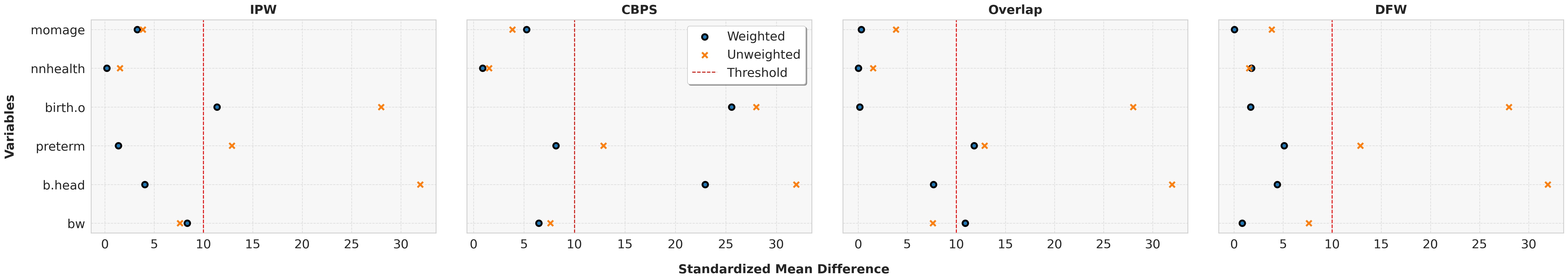}
	\caption{Absolute standardized mean difference (smaller the better) on the IHDP dataset.}
	\label{fig:4}
\end{figure*}
\begin{figure*}[!htb]
	\centering
	\setcounter{figure}{4}
	\includegraphics[width=1\textwidth]{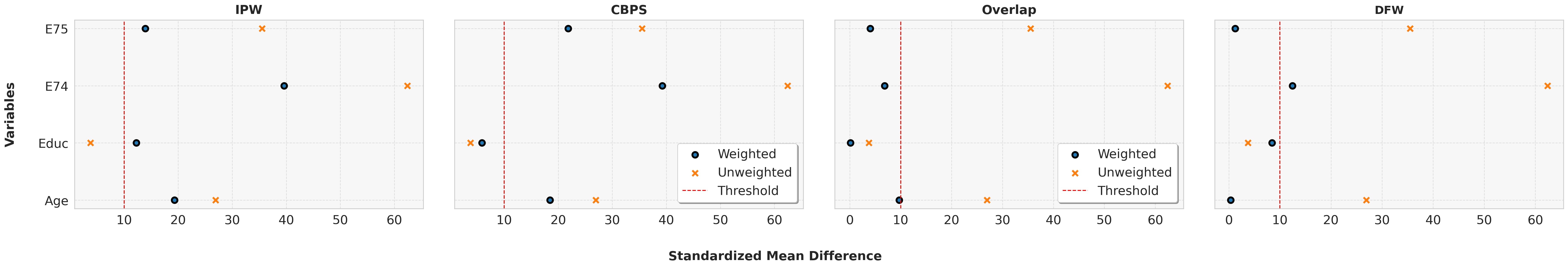}
	\caption{Absolute standardized mean difference (smaller the better) on Jobs.}
	\label{fig:5}
\end{figure*}

\begin{figure*}[!htb]
	\centering
	\captionsetup{justification=centering}
	
	\setcounter{figure}{5}
	\includegraphics[width=1\textwidth]{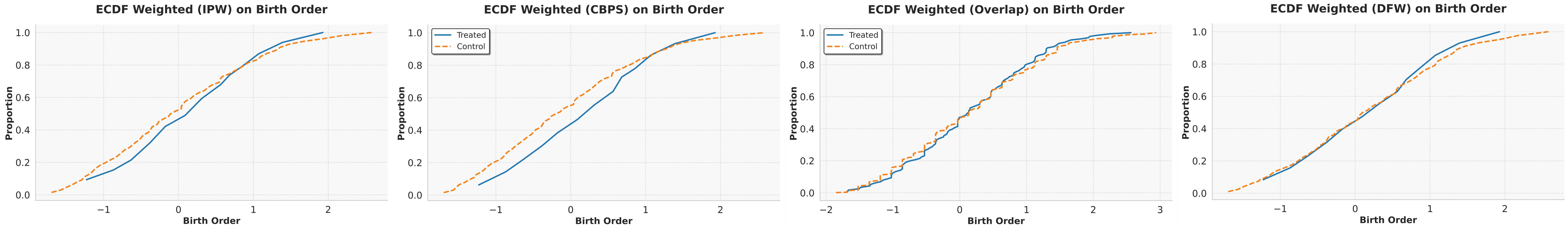}
	
	\caption{ECDF plots show distribution identicality between the treated and control groups on the birth order covariate (IHDP).}
	\label{fig:6}
	
\end{figure*}

\begin{figure*}[!htb]
	\centering
	\captionsetup{justification=centering}
	
	\setcounter{figure}{6}
	\includegraphics[width=1\textwidth]{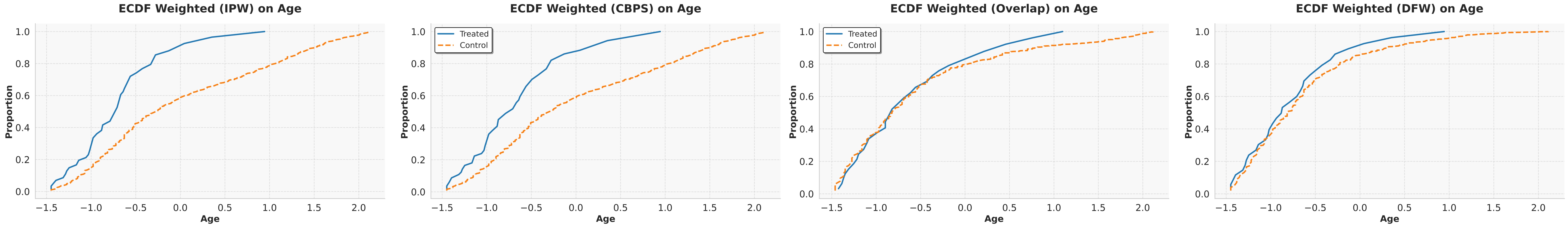}

	\caption{ECDF plots show distribution identicality between the treated and control groups on Age (Jobs).}
	\label{fig:7}
	
\end{figure*}

\begin{figure*}[!htb]
	\centering
	\captionsetup{justification=centering}
	
	\setcounter{figure}{7}
	
	\includegraphics[width=1\textwidth]{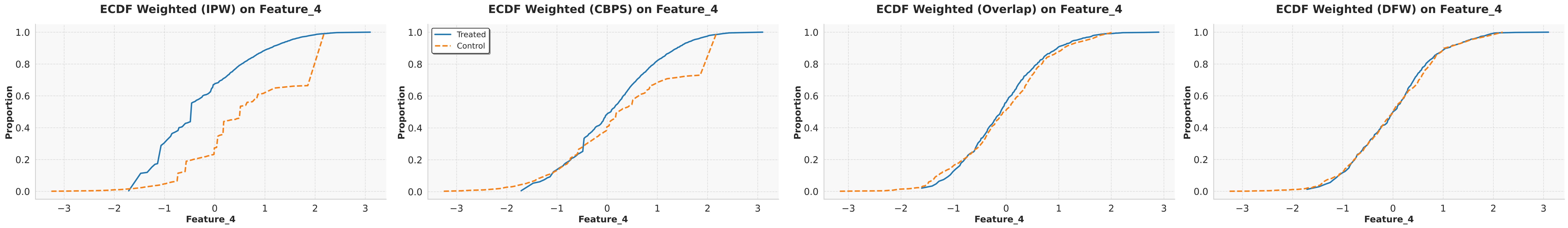}
	
	\caption{ECDF plots show distribution identicality between the treated and control groups on feature no. 4 (highly biased synthetic dataset).}
	\label{fig:8}
	
\end{figure*}

\begin{figure}[!htb]
	\centering
	\setcounter{figure}{8}
	\includegraphics[width=0.45\textwidth]{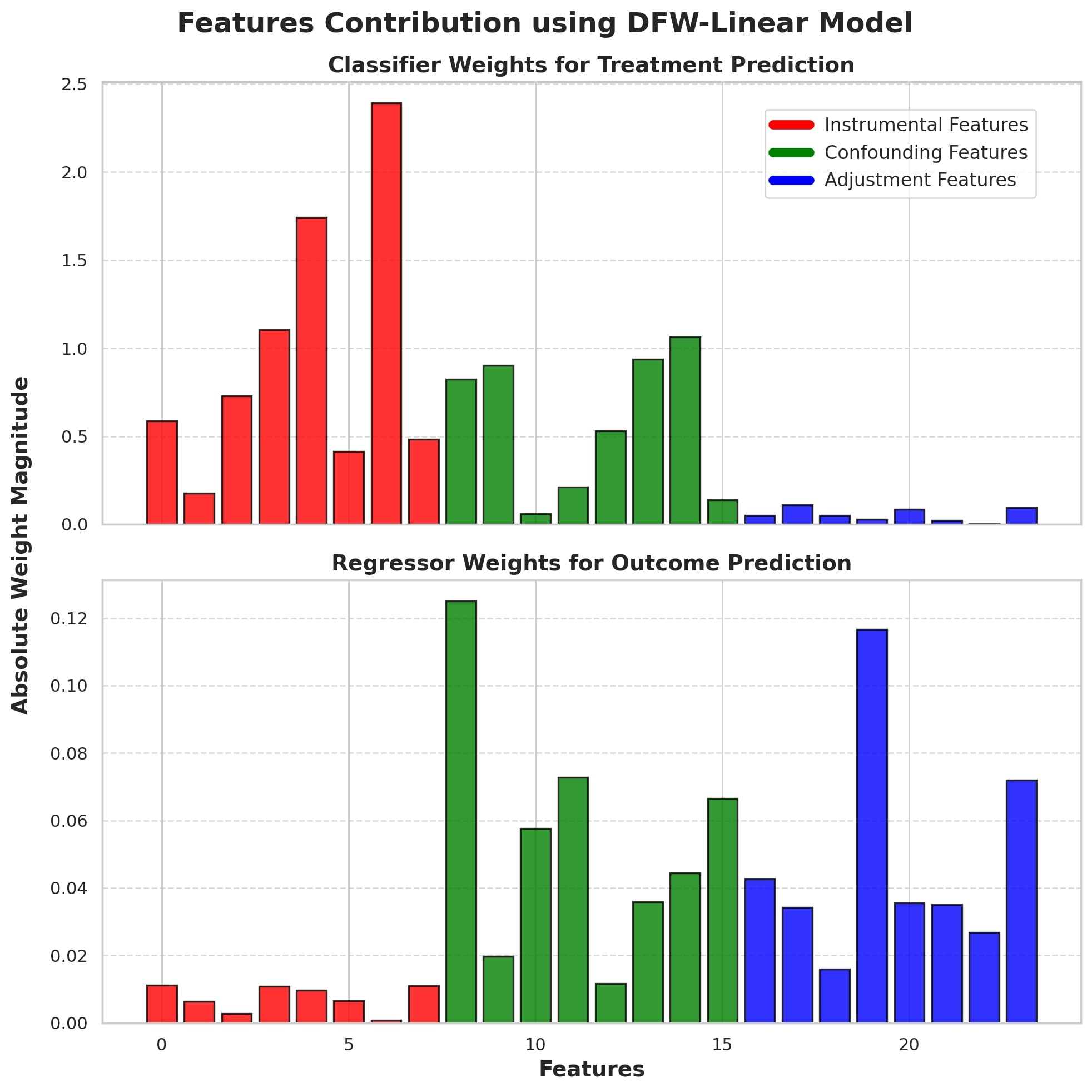}
	\caption{We validate our method on a synthetic dataset {\protect\cite{Khan2024OnTE}} where the feature roles are predefined: instrumental features influence only treatment assignment, confounding features affect both treatment assignment and outcome, and adjustment features impact only the outcome. For our weighting scheme to function as intended, the classifier should rely exclusively on instrumental and confounding features for propensity score estimation. More importantly, the regression model should leverage only confounding and adjustment features for outcome prediction using DFW weights. The figure confirms these expected patterns, demonstrating that our approach effectively separates feature contributions as intended.}
	\label{fig:9}
\end{figure}

\end{document}